\documentclass[10pt,twocolumn]{article}

\usepackage{cogsci}

\cogscifinalcopy 

\usepackage{pslatex}
\usepackage{apacite}
\usepackage{float} 

\usepackage[utf8]{inputenc} 
\usepackage[T1]{fontenc}    
\usepackage{url}            
\usepackage{booktabs}       
\usepackage{amsfonts}       
\usepackage{nicefrac}       
\usepackage{microtype}      
\usepackage[dvipsnames]{xcolor}
\usepackage{tabularx}
\usepackage{graphicx}
\usepackage{hyperref}       
\usepackage{titlesec}
\titlespacing*{\subsubsection}{0pt}{1ex}{1ex}

\usepackage[round, sort]{natbib}
\usepackage{xcolor}

\newcommand{\gtlogo}{\raisebox{0pt}{\includegraphics[scale=0.04]{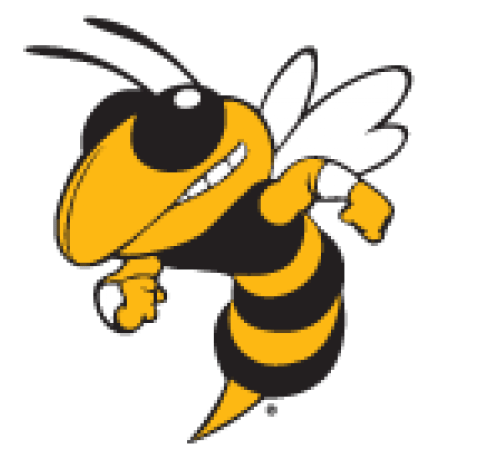}}}

\title{A Neural Network Model of Complementary Learning Systems: Pattern Separation and Completion for Continual Learning}

\author{{\large \bf James P Jun, Vijay Marupudi, Raj Sanjay Shah, Sashank Varma} \\
  \{jjun44, vijaymarupudi, rajsanjayshah,
varma\}@gatech.edu
  \\
  Georgia Institute of Technology \gtlogo
  }

\begin{document}

\maketitle

\begin{abstract}

Learning new information without forgetting prior knowledge is central to human intelligence. In contrast, neural network models suffer from catastrophic forgetting: a significant degradation in performance on previously learned tasks when acquiring new information. The Complementary Learning Systems (CLS) theory offers an explanation for this human ability, proposing that the brain has distinct systems for pattern separation (encoding distinct memories) and pattern completion (retrieving complete memories from partial cues). To capture these complementary functions, we leverage the representational generalization capabilities of variational autoencoders (VAEs) and the robust memory storage properties of Modern Hopfield networks (MHNs), combining them into a neurally plausible continual learning model.  We evaluate this model on the Split-MNIST task, a popular continual learning benchmark, and achieve close to state-of-the-art accuracy (\textasciitilde90\%), substantially reducing forgetting. Representational analyses empirically confirm the functional dissociation: the VAE underwrites pattern completion, while the MHN drives pattern separation. By capturing pattern separation and completion in scalable architectures, our work provides a functional template for modeling memory consolidation, generalization, and continual learning in both biological and artificial systems.


    

\textbf{Keywords:}
catastrophic forgetting; catastrophic interference; continual learning; complementary learning systems.
\end{abstract}

\section{Introduction}
While neural networks excel at learning new tasks, they often suffer from \emph{catastrophic forgetting} (CF), where performance on previously learned tasks drops precipitously when learning a new task \citep{McCloskey1989}. In this regard, they differ from humans, for whom \emph{continual learning} (CL) is natural \citep{Mitchell2018, Parisi2019}. In artificial systems, CL is typically supported through direct rehearsal \citep{rolnick2019experience, gandhi2024natural}, involving the replay of stored examples, or through synthetic rehearsal where generative models recreate prior experiences. Among synthetic rehearsal approaches, \emph{generative replay} (GR), which interleaves synthetic samples of previous tasks with new training data, has proven to be highly effective in mitigating catastrophic forgetting \citep{vandeven2020, vandeven2022}. GR strategies draw direct inspiration from the CLS theory of the mammalian brain, which posits distinct but interacting memory systems within the neocortex and hippocampus \citep{Mcclelland1995, OReilly2014, Kumaran2016}.


Over time, the pursuit of state-of-the-art performance on CL tasks has detached GR approaches from their neural inspiration, sacrificing neural plausibility for engineering gains. Here, we describe a model architecture guided by the functional principles of the CLS framework. Rather than replicating the \emph{structures} of the hippocampus, as computational cognitive neuroscientists have previously done \citep{Mcclelland1995, Norman2003, Schapiro2017}, we abstract to the \emph{functions} performed by those structures in concert with the cortex: \emph{pattern separation} and \emph{pattern completion} \citep{Bakker2008, Larocque2013}. Building on the biologically plausible architecture proposed by \citet{Spens2024}, we extend it to a continual learning application by integrating the hippocampal (Modern Hopfield Network) and neocortical (variational auto-encoder) components. In consolidating these into a unified architecture (VAE+MHN model), we advance both performance and neuro-plausibility of this framework by enabling simulation of dynamic neocortical-hippocampal interactions during memory encoding and retrieval. We evaluate the model on the standard Split-MNIST benchmark and systematically track both pattern separation and pattern completion across continual learning, using quantitative representational analyses, Euclidean distances for separation, and structural similarity index measure (SSIM) scores for completion, to validate functional dissociations. Finally, we demonstrate that the model achieves high continual learning performance similar to non-continual learning baselines while preserving clear functional roles aligned with CLS principles.




\subsection{The CLS Framework}

\begin{figure*}[t]
    \centering
    \includegraphics[width=0.88\textwidth]{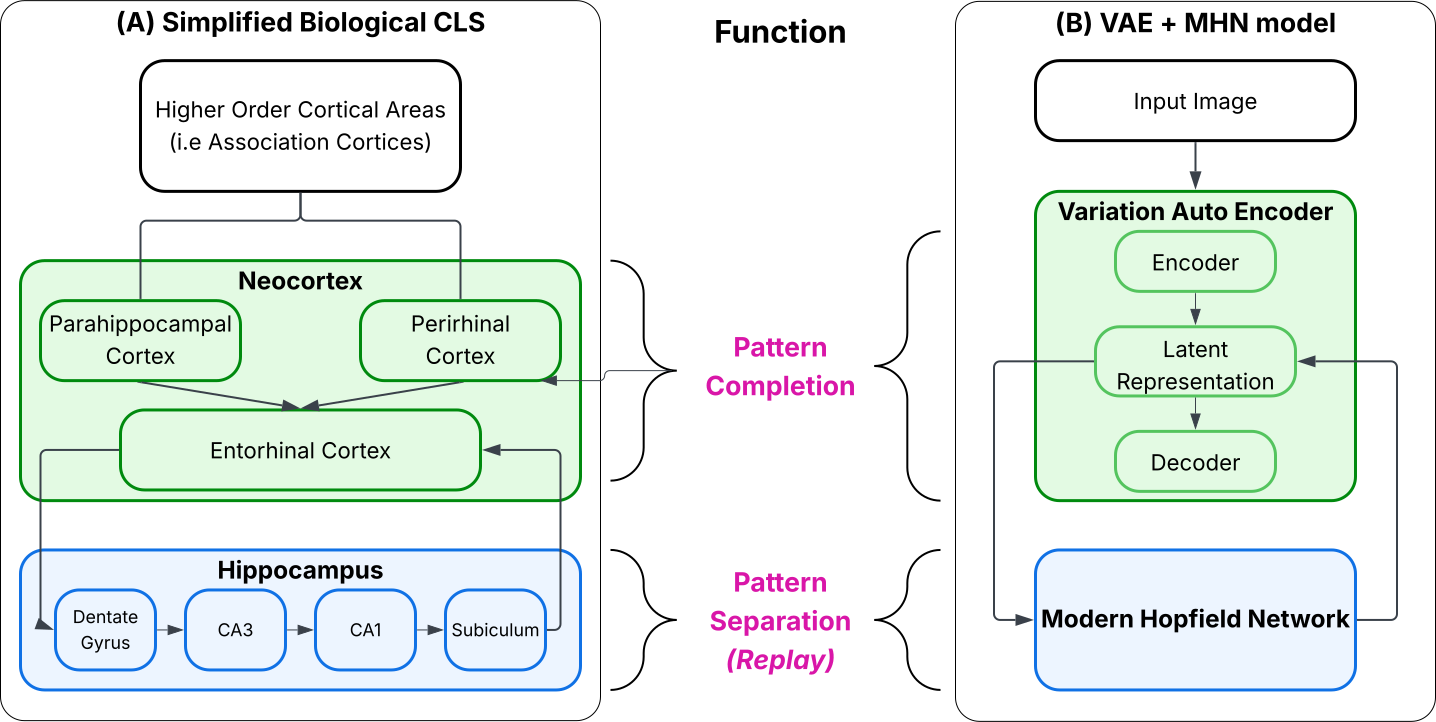}
    \caption{(A) Schematic of the CLS framework, showing the division of memory functions between the hippocampus, which supports pattern separation, and the neocortex, which supports pattern completion. (B) Computational instantiation of these functions in the VAE+MHN model: the MHN enables pattern separation (blue) by storing distinct episodic representations, while the VAE enables pattern completion (green) by abstracting generalized representations from replayed memories. The colors highlight the direct functional mapping between biological structures and their computational counterparts.}
    
    \label{fig:mtl}
\end{figure*}

In humans, the CLS framework directly addresses the question at the
heart of CL: How does the brain learn new information while retaining
older memories \citep{Kumaran2016, Mcclelland1995, OReilly2014}? CLS
posits that this is achieved through two complementary systems: (1)
the hippocampus, which encodes episodic memories or the detailed
snapshots of ongoing experiences, and (2) the neocortex, which
gradually abstracts statistical regularities to form semantic
memories. The hippocampus primarily supports \emph{pattern
separation}, encoding similar experiences as distinct representations
to minimize interference. To facilitate consolidation, it actively
\emph{replays} episodic memories over time, training the
slower-learning neocortex. The neocortex, in turn, specializes in
\emph{pattern completion}: reconstructing complete memories from
partial or degraded inputs. Figure~\ref{fig:mtl}A illustrates these
core structures and their roles within the CLS framework.



\subsection{Neuroscience-Inspired Generative Replay Methods for CL}

Early approaches to CL in artificial systems often relied on
\emph{explicit replay}, wherein models directly stored and replayed
examples from prior tasks \cite{rolnick2019experience}. While
effective, explicit replay poses serious limitations: it often
requires storing large amounts of past data \cite{gandhi2024natural},
scales poorly with the number of tasks \cite{krawczyk2024analysis},
and lacks biological plausibility, as brains do not store exact copies
of past experiences. To address these shortcomings, machine learning
researchers turned to \emph{synthetic replay} methods inspired by the
CLS framework and, in particular, the hippocampal mechanism of memory
replay. Standard GR implements this strategy by training a separate
model to generate prior task samples alongside new task data. More
recently, Brain-Inspired Replay (BI-R) refines this approach by
generating compressed latent representations rather than raw inputs,
further improving scalability and achieving strong performance on
continual learning benchmarks such as Split-MNIST \citep{vandeven2020,
  vandeven2022}.


Through these advances, synthetic replay methods have come to diverge from core principles of biological memory systems. For instance, BI-R saves a separate copy of the generator after each task, an approach that is computationally inefficient and biologically implausible. 
Recent work by \citet{Spens2024} increases biological plausibility by pairing a VAE with an MHN \citep{ramsauer2021hopfieldnetworksneed}, where episodic memory replay from the MHN supports gradual learning in the VAE (Figure~\ref{fig:mtl}B). Replay from this network serves as a form of ``teacher-student learning'' \citep{Hinton2015} to train the VAE. The VAE, in turn, gradually extracts relevant features from replayed samples from the MHN to learn a generalized version of the images presented. However, while their VAE+MHN model demonstrates pattern completion on a static MNIST task, these researchers did not test for pattern separation nor did they evaluate performance under continual, sequential learning settings like Split-MNIST. 

In contrast, our model departs from prior work in three key ways: (1) adaption to continual learning to model human learning without storing separate, task-specific models, thereby preserving biological plausibility; (2) implementing latent-space replay instead of raw image replay to align with scalability and neuro-plausibility; and (3) introducing functional analyses to track pattern separation and completion over time, enabling direct evaluation of CLS principles for continual learning.


\subsection{The Current Study}

In this work, we extend the VAE+MHN model to explicitly operationalize both CLS functions: pattern separation and pattern completion, under a continual learning setting. We evaluate the model on the Split-MNIST benchmark, a popular class-incremental task in the machine learning literature. Our results show that the model achieves continual learning performance close to an upper baseline model trained non-continually, without the need for explicit task labels or stored replay buffers. This work demonstrates that biologically inspired memory mechanisms can support continual learning in neural networks.


\begin{figure}[h]
  \centering
  \includegraphics[width=0.38\textwidth]{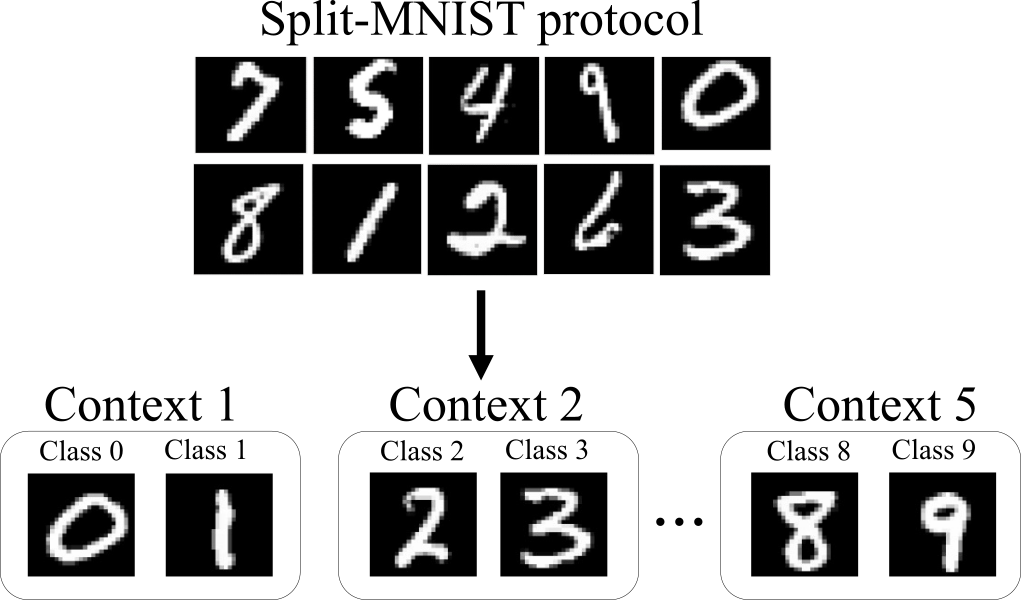}
  \caption{Class-incremental continual learning setup on the MNIST dataset (Split-MNIST Protocol).}\label{fig:split-mnist}
\end{figure}

\section{Method}

The VAE+MHN model was trained on the Split-MNIST task (Figure~\ref{fig:split-mnist}), a benchmark commonly used in continual learning research. In the class-incremental setting, the MNIST dataset (60,000 training and 10,000 test images of handwritten digits, 0–9) is divided into five sequential binary classification tasks, each involving approximately 12,000 images from two classes (e.g., digits 4 and 5). The model experiences each new task \textit{incrementally}, encountering only two new classes at each stage. 
This setup simulates continual learning by requiring the model to acquire new classes while retaining prior classes.

To understand how architectural hyperparameters impact memory retention and generalization, we systematically explored several model parameters. First, we tested if pattern completion affects eventual memory consolidation by comparing the latent layer (which supports pattern completion) or the fully connected layer (which performs a transformation but lacks generative properties) as inputs to the MHN. We also varied the dimensionality of the latent layer and the replay rate (i.e. ratio of replayed samples from older contexts to new context samples) during training and evaluated its impact on image reconstruction.

Training followed a generative replay procedure aligned with CLS principles. During the first context 1 (0 vs. 1), the model learned from real images without replay, as no prior memories existed. At the end of stage 1 training, 5\% of the images (\textasciitilde600 samples) were randomly selected, and their latent representations from the VAE were stored in the MHN. During each subsequent stage, the model was trained on two types of samples: (1) from the current stage and (2) MHN generated samples representing those from prior stages. To produce replay samples, random noise vectors were used as retrieval cues. The MHN performed recurrent settling by iteratively updating its state to minimize an internal energy function \citep{ramsauer2021hopfieldnetworksneed}. The settled MHN state was ``replayed'' (i.e., decoded through the VAE’s decoder), generating synthetic replay samples corresponding to previously learned classes. These replayed samples were interleaved with the new class data during training with the goal of mitigating forgetting.



\begin{figure}[t]
  \hspace{20pt}
  \centering
  \includegraphics[width=0.5\textwidth]{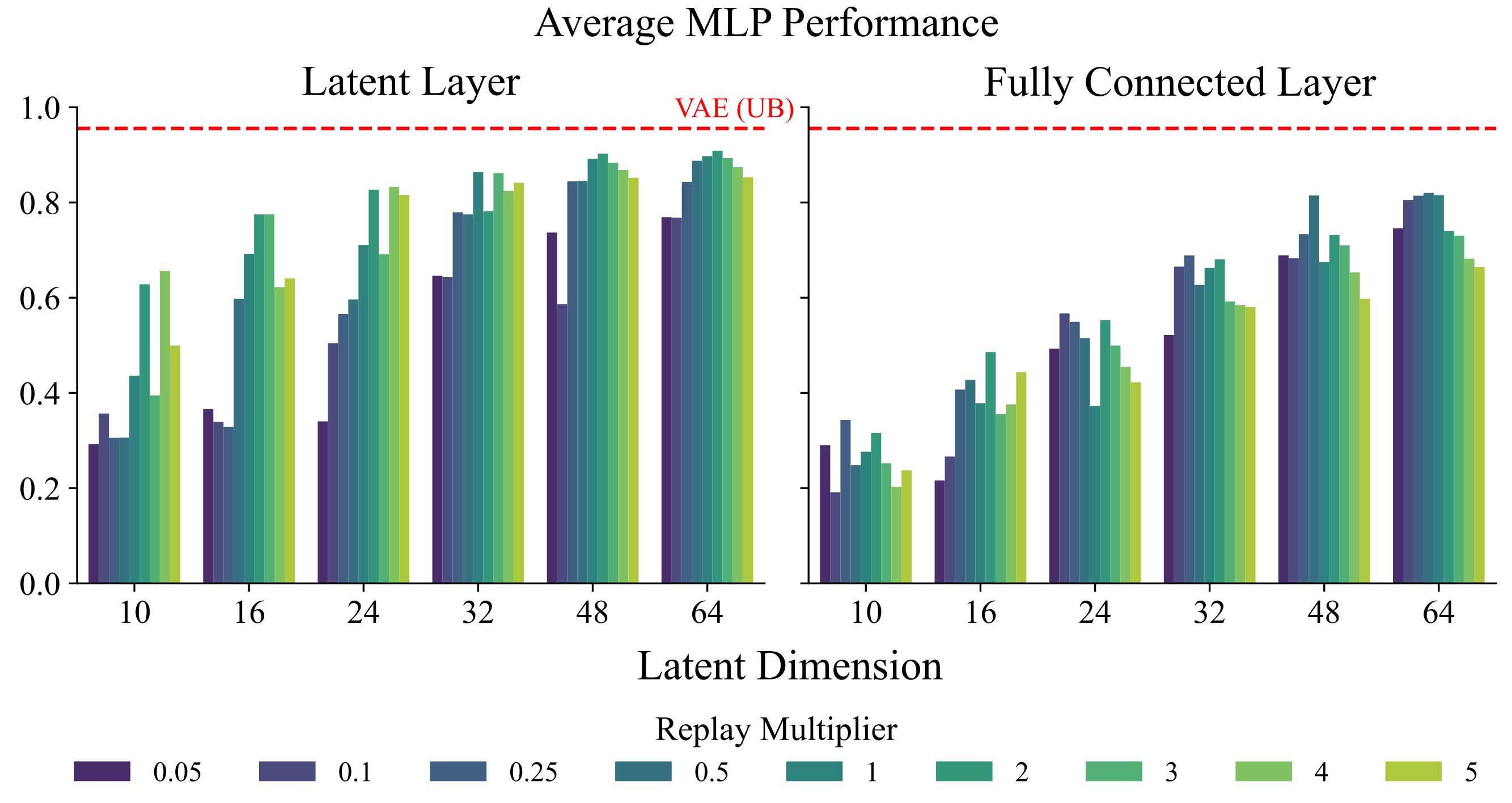}\
  \caption{Different parameters were evaluated using Split-MNIST to determine the optimal model architecture. Each experiment was repeated 5 times; the average results are plotted. Upper baseline (VAE UB) performance is shown in red.}\label{fig:model-architecture-explorations}
\end{figure}

As the generative model operates in an unsupervised manner, evaluating its performance requires judging whether the images it generates are valid numbers. To do so, we trained a separate MLP classifier on MNIST (achieving 97.58\% test accuracy) and used it to classify images that were reconstructed by the model (demonstrated in Fig~\ref{fig:model-architecture-explorations}). Performance was assessed against the corresponding class label $0-9$.
 
All VAE models were trained for 10 epochs with a batch size of 32 and a fixed learning rate of 1E-3 with images normalized between 0 and 1 and shuffled to prevent order bias. A VAE loss function combining binary cross-entropy (for reconstruction accuracy) and Kullback-Leibler divergence (to regularize the latent space), with losses from current and (when applicable) replayed images, was summed and backpropagated to update model weights. For the MHN, a large inverse temperature parameter ($\beta$) was selected to ensure individual memory recall.

\section{Results}


\subsection{Model performance}


Model performance varied substantially with different hyperparameter settings (see Figure~\ref{fig:model-architecture-explorations}). Storing the latent layer in the MHN yielded better image reconstruction, particularly as latent dimensionality increased. Replay rate showed a quadratic effect: both low and high ratios impaired performance, with optimal results near a 1:1 replay ratio. We selected the model that stored latent layer representations with a dimensionality of 48 and used a 1:1 replay ratio for reasons of parsimony. This model achieved a test accuracy of 89.71\%, which is
just below the most accurate model (91.01\%).
All further results reported in this paper refer to this selected model. The performance of this model trained on Split-MNIST and three useful comparison models are reported in Table~\ref{tab:accuracy-of-model}. To contextualize performance, we compared the selected VAE+MHN model to three benchmarks (Table~\ref{tab:accuracy-of-model}). The upper baseline (UB) model, which was a VAE trained on all 10 MNIST classes simultaneously, reached 95.55\% accuracy with no CF, as expected. The lower baseline (LB) was an untrained VAE. The control model, a standard VAE trained sequentially without GR or MHN, exhibited severe CF, achieved test accuracy nearly 30\% less than the UB. In contrast, the VAE+MHN model successfully mitigated CF, achieving test accuracy within 6\% of the UB.

\subsection{VAE+MHN Performance in Sequential Learning}
  \vspace{-15pt}
\begin{table}[h]
  \caption{The test accuracy of the VAE + MHN model in the CL setting approached the performance of the upper baseline (UB). Offline: Offline learning; CIL: Class Incremental Learning.}
  \label{tab:accuracy-of-model}
\begin{tabularx}{0.5\textwidth}{X X r}
  \toprule
  Model & Training  & Performance \\
    &  paradigm &  (via MLP) \\
  \midrule
  VAE (UB) & Offline & 95.55\% \\
  VAE (LB) & Untrained & 8.92\% \\
  VAE (Control) & CIL & 67.75\% \\
  \textbf{VAE+MHN} & CIL & \textbf{89.71\%} \\
  \bottomrule  
\end{tabularx}
\end{table}

\subsection{Pattern Separation vs Pattern Completion}

Neuroscience studies have mapped the roles of the hippocampus and neocortex in pattern separation and pattern completion, respectively \citep{Mcclelland1995, Norman2003, OReilly2014, Kumaran2016, Schapiro2017}. The current study explored how these functions emerge in the VAE+MHN model during continual learning. Guided by the CLS framework, we hypothesized that the MHN would primarily align with hippocampal processing and the pattern separation function. In contrast, the VAE's latent space would align with neocortical processing and the pattern completion function. We analyzed the latent space of the VAE and MHN components of the VAE+MHN model for evidence of these predictions. As references, we also considered the latent spaces of the UB VAE and Control VAE models. The former was trained on all classes simultaneously and should therefore have learned an optimized representation of the relationships between all classes. By contrast, the latter model was trained in a CL setting and experienced catastrophic forgetting, and should have learned an impoverished and incomplete representation.  

\subsubsection{Pattern separation}

We quantified intra-class pattern separation. To do this, a set of 128 held-out images per MNIST class was passed through the VAE+MHN (CL) model trained on SplitMNIST, an UB VAE classically trained on MNIST, and a Control VAE trained on Split-MNIST without replay. Representations were extracted at both the VAE latent layer and MHN (if present). MHN representations were extracted by passing latent layer activations of the test image as retrieval cues. All pairwise Euclidean
distances within each class were computed and averaged. The larger this distance, the more distinct the representations, and thus the greater the pattern separation.

\begin{figure}
  \centering
  \includegraphics[width=0.4\textwidth]{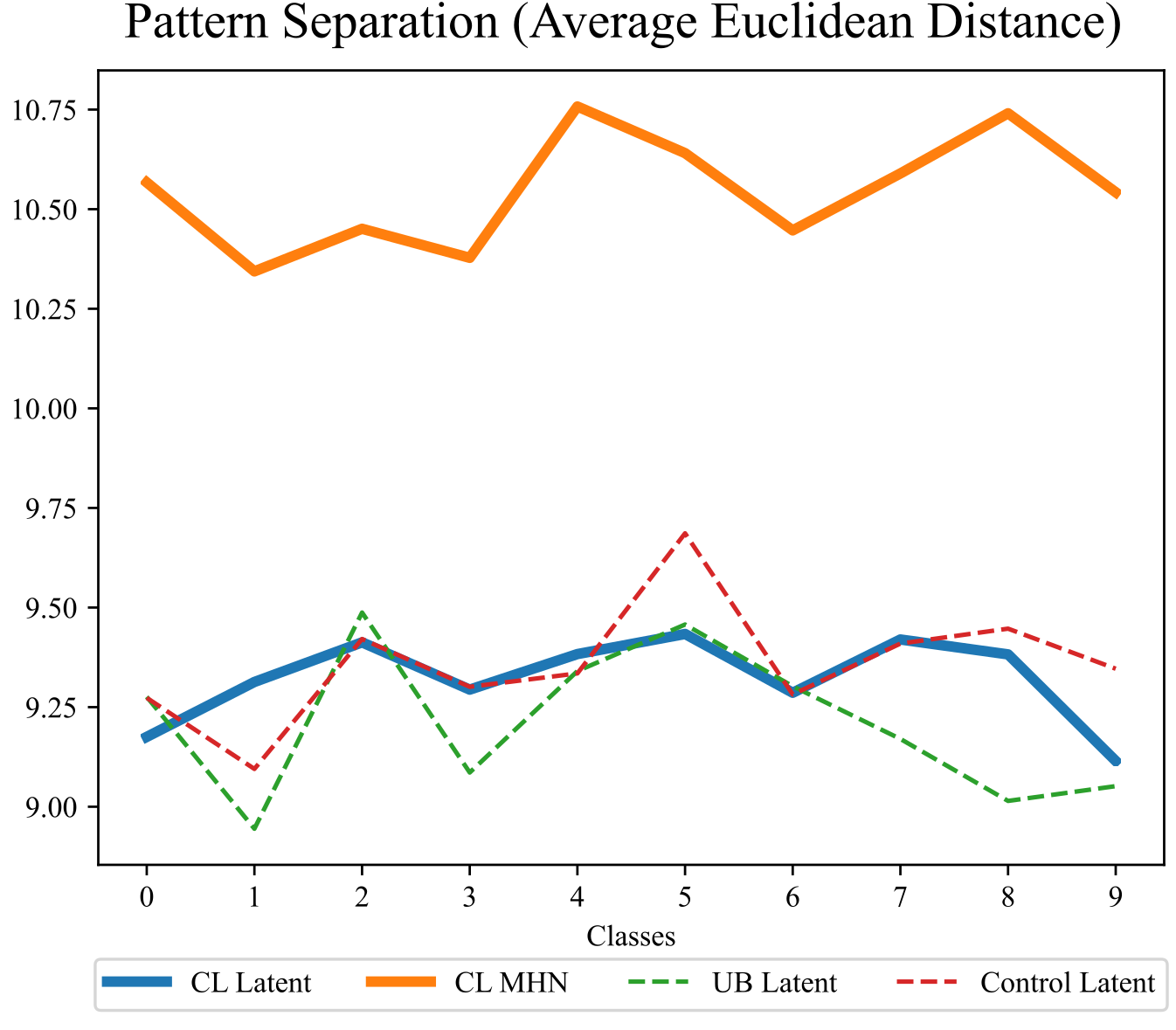}
  \caption{Averaged intra-class Euclidean distance across layers/models is shown. Consistent with CLS, representations within the MHN showing greater pattern separation (i.e., greater Euclidean distance between samples within a class).}\label{fig:pattern_separation}
\end{figure}

Pattern separation on the VAE’s latent layer (i.e., neocortex) and the MHN (i.e., hippocampus) is shown in Figure~\ref{fig:pattern_separation}. As predicted by the CLS, the MHN has learned the more differentiated representation necessary for supporting the pattern separation function, with a large Euclidean distance between the representations of images of the same class. By contrast, the VAE's latent layer shows smaller Euclidean distance between representations of images of the same class, i.e., they are more similar to (and less differentiated from) each other.

This difference is further illustrated in Figure~\ref{fig:t-SNE}, which shows the t-distributed stochastic neighbor embedding (t-SNE) of images of the same class, color coded, as the 5 contexts are learned. The high dimensionality of the activation vector representations is reduced to visualize their underlying structure. As the VAE+MHN model sequentially learns new classes, the MHN representations are more sparsely scattered across the space, reducing overlap between images of the same class and supporting pattern separation. By contrast, the VAE's latent layer representations reveal clear clusters by class. This supports pattern completion and is antithetical to pattern separation (although, as we will see next, these clustered representations support pattern completion and generalization). Thus, the different representations of the two components align with CLS expectations.


To assess statistical significance, we compared intra-class Euclidean distances of the VAE+MHN model's MHN representation and the VAE latent layer representations from the same model, the UB VAE, and the Control VAE. For each MNIST class, we computed all pairwise Euclidean distances between layer representations within that class and performed a two-sample t-test comparing thee MHN representation to each comparator model. This resulted in 30 total comparisons (10 classes x 3 layer pairs), for which we applied Bonferroni correction to control for family-wise error. In all cases, the MHN
representation exhibited significantly greater intra-class distance (Bonferroni-corrected $p <0.001$ after correction), indicating the  MHN representation shows enhanced pattern separation within the representational space.



\begin{figure}
  \vspace{-20pt}
  \centering
  \includegraphics[width=0.3\textwidth]{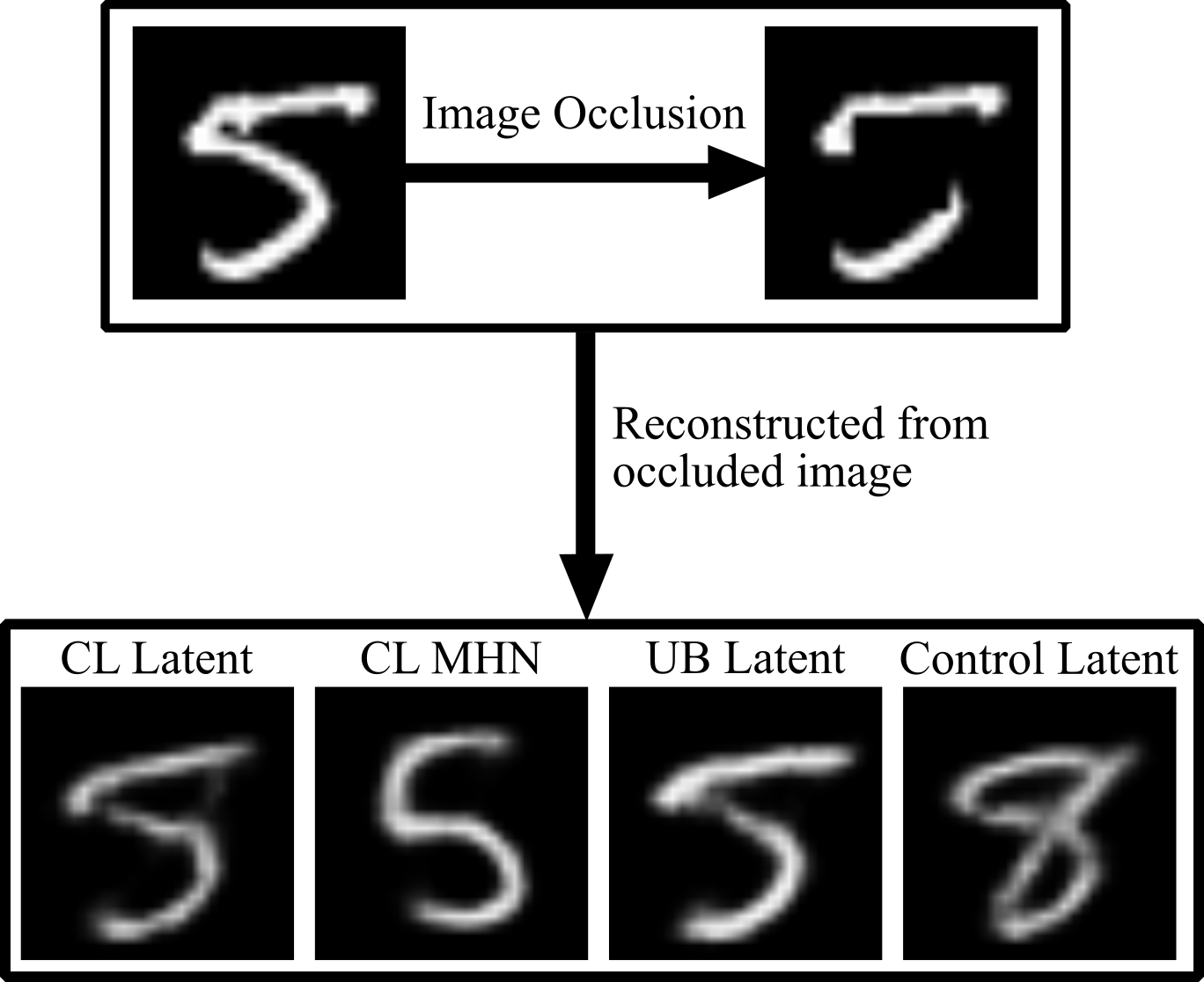}
  \caption{Held-out test images are perturbed (e.g., center occluded) and input into the VAE+MHN model, the UB VAE model, and the Control VAE model. Perturbed images are encoded and then decoded to reconstruct the original image for use in quantifying pattern completion.}\label{fig:pattern-completion-reconstruct}
\end{figure}

\subsubsection{Pattern completion}

To measure pattern completion, we again analyzed the VAE latent layer and MHN representations.
We perturbed test images via center occlusion by adding a 10x10 pixel black mask. Each perturbed image was processed by the VAE+MHN model until the latent layer. First, the latent representation was directly decoded to reconstruct an unperturbed image. Second, the latent representation was projected to the MHN to retrieve a hippocampal representation, which was then decoded to reconstruct an unperturbed image. This process is depicted in Figure~\ref{fig:pattern-completion-reconstruct}. For comparison, the perturbed image was also presented to the UB and Control models and reconstructed. The similarity of each of the four reconstructed images to the original, unperturbed image was then computed using the Structural Similarity Index Measure (SSIM) \citep{Wang2004}. This is a perceptual measure that captures the structure, contrast, and luminance similarity between a reference image and a reconstructed image. The greater the SSIM, the better the reconstruction, and thus the more effective the pattern completion. 

\begin{figure}
  \vspace{-10pt}
  \centering
  \includegraphics[width=0.4\textwidth]{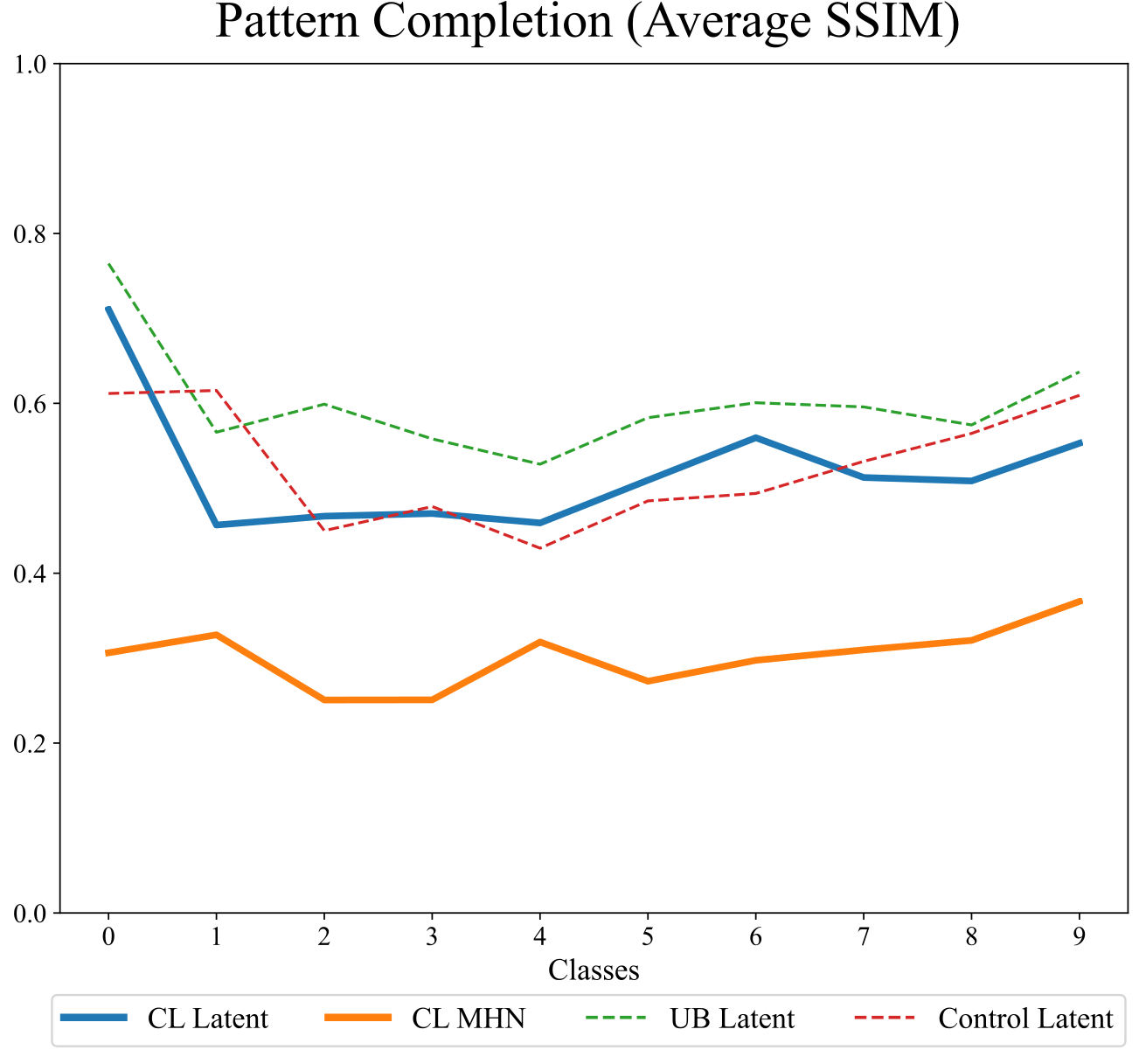}
  \caption{Structural Similarity Index Measure (SSIM) scores across models between reconstructed occluded stimuli and original, unperturbed stimuli are shown. Consistent with CLS, VAE latent layer representations support pattern completion.}\label{fig:pattern-completion-ssim}
\end{figure}

The results are shown in Figure~\ref{fig:pattern-completion-ssim}. For each of the 10 classes, reconstruction from the VAE+MHN model’s VAE latent layer representations resulted in greater similarity to the original, unperturbed image than reconstruction from the MHN representations. In fact, reconstruction from the VAE's latent layer resulted in a performance close to that of the UB.

\begin{figure}
  \centering
  \includegraphics[width=0.45\textwidth]{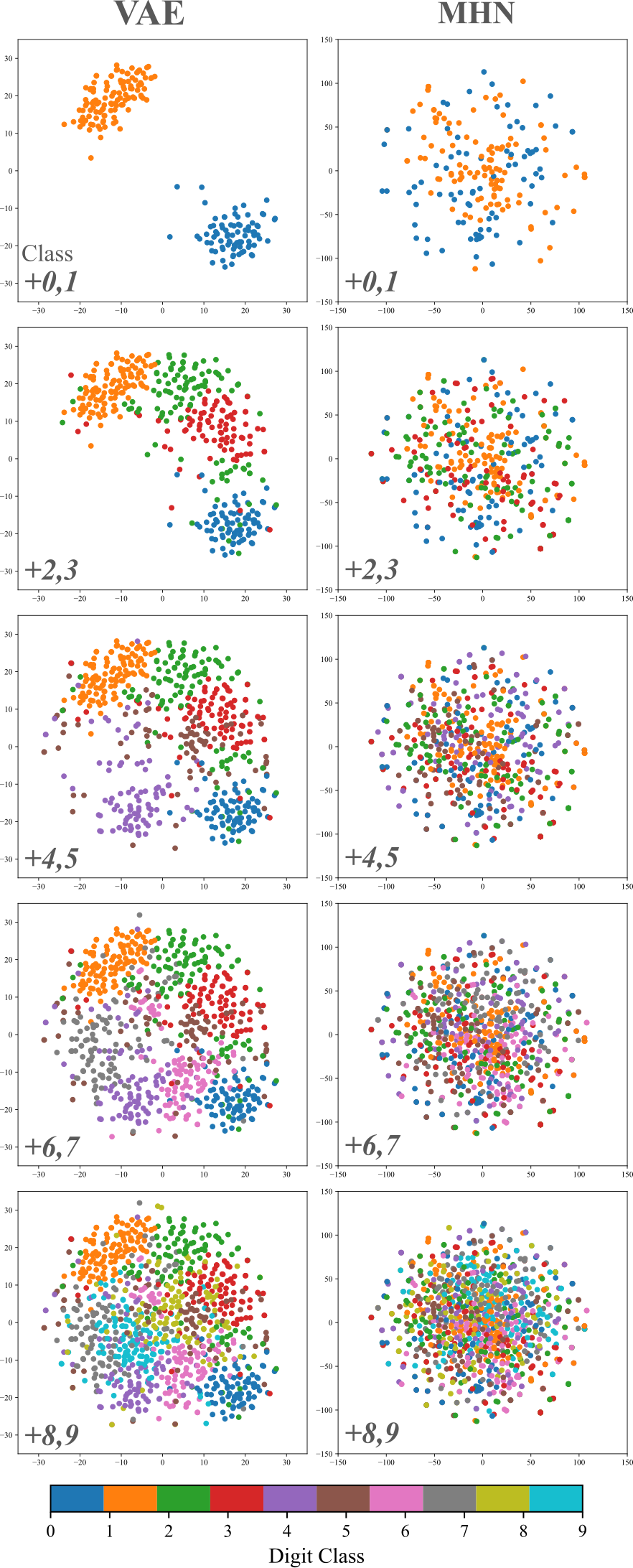}
  \caption{Visualization of t-SNE projections of latent representations from the VAE latent layer (left) and MHN (right) for classes 0-9. Each panel depicts the latent space as the number of classes increases during CL. The VAE latent layer exhibits clustering of the images of each class, supporting its role in pattern completion. By contrast, the MHN shows sparser and more distributed representations of images within each class, supporting its role in pattern separation.}\label{fig:t-SNE}
\end{figure}

Referring to Figure~\ref{fig:t-SNE}, we see that the VAE's latent layer representation groups images of the same class. This clustering supports the pattern completion function. The VAE's latent layer representation efficiently consolidates class-level features despite perturbation. By contrast, the  MHN prioritizes differentiation over consolidation, making it less suited for the reconstruction of perturbed images. This again reinforces the complementary roles of these components within the VAE+MHN model, consistent with the CLS framework.

We conducted statistical comparisons of intra-class SSIM values across models layers, using the VAE+MHN model's MHN representation as the baseline. Specifically, we compared the SSIM values between the MHN  of this model's VAE latent layer representation and the VAE latent layer representations of the UB and Control models. For each MNIST class, we performed a two-sample t-test comparing the MHN's SSIM values against each comparator. Bonferroni correction was applied across all 30 comparisons. In all cases, the MHN representation exhibited lower SSIM values (Bonferroni-corrected $p<0.001$ after correction), indicating poorer reconstruction consistent with its role in differentiating similar inputs in representational space. 


\section{Discussion}

We evaluated whether the \citet{Spens2024} model, inspired by the CLS framework \citep{Mcclelland1995, Norman2003, OReilly2014, Kumaran2016, Schapiro2017}, is able to overcome CF and demonstrate CL. We trained this VAE+MHN model (and several baseline models) on Split-MNIST, a popular class-incremental continual learning benchmark.
We applied the concept of teacher-student learning using a fast, hippocampal learner as a “teacher” (MHN) of previous classes to a slow, neocortical “student” (VAE).
The model incrementally learned all of the classes and showed high test accuracy at the end of training (89.71\%), within range of the upper baseline VAE trained on all classes simultaneously (95.55\%). Critically, a representational analysis showed that the MHN component (i.e., the hippocampus analog) supports pattern separation and the VAE’s latent layer (i.e., the neocortex analog) supports pattern completion. These findings align with the CLS theory.

Unlike ML models, which focus on the engineering goal of overcoming CF and achieving CL (e.g., \citet{vandeven2022}), the VAE+MHN model places a higher priority on fidelity to the memory architecture of the mammalian brain. Whilst not a 1:1 mapping, MHNs appear to capture key qualities of the hippocampus, particularly in modeling associative memory via CA3 pyramidal neurons and recurrent connectivity \citep{Krotov2021}. Our findings support their neural plausibility in showing that they capture the pattern separation function attributed to the hippocampus's CA3 subfield. Also critical from the perspective of neural plausibility is that the MHN is capable of being incrementally updated during learning. This captures the high-frequency interaction between the hippocampus and neocortex during learning, and differentiates the VAE+MHN model from ML approaches that are also inspired by the CLS framework. For example, BI-R saves a separate copy of the generative model (used to generate replayed samples from old tasks) after learning each new task, which is not neurally plausible \citep{vandeven2020, vandeven2022}. That said, the VAE+MHN model greatly simplifies the biological interactions within the brain compared to more biologically inspired CLS models \citep{Mcclelland1995, Norman2003, Kumaran2016, Schapiro2017}, which model the individual subfields of the hippocampus and their connectivity to each other. On the other hand, these models are difficult to scale from laboratory tasks to more demanding ML tasks. The VAE+MHN model sits between these two extremes and provides an opportunity to bridge insights from both computational neuroscience and scalable model architectures.

Our findings should be considered in the context of several limitations. From the cognitive neuroscience perspective, the VAE+MHN model’s design simplifies many complex interactions between and within the neocortex and hippocampus of the mammalian brain. Many other structures shape memory, such as the amygdala, prefrontal cortex, cerebellum, basal ganglia, etc. More generally, the model does not fully capture the dynamic and interdependent nature of neural processing.

Our analyses suggest complementary roles for the VAE and MHN in supporting pattern separation and completion, opening up multiple avenues for future exploration. One promising direction is to investigate the neural plausibility of this model beyond basic memory functions. The medial temporal lobe also supports higher-order cognitive processes such as transitive inference, paired associative inference, and acquired equivalence paradigms \citep{Kumaran2012, Wendelken2010, Gross2007}. While simpler, more neurally aligned models have explored the implementation of these functions \citep{Schapiro2017}, it remains an open question whether biologically grounded mechanisms like those in the VAE+MHN architecture can extend to such forms of reasoning. For example, when trained on "A is greater than B" and "B is greater than C", will the model understand that "A is greater than C", and if so, which components will underwrite that understanding? Our findings confirm that the MHN and VAE components implement dissociable mechanisms of pattern separation and completion, respectively, core functions implicated in memory consolidation. Investigating whether these mechanisms can also support higher-order processes like transitive inference offers a promising direction for understanding how memory structures contribute to reasoning and generalization.

\bibliography{CogSci_Template.bib}
\end{document}